\documentclass[journal]{IEEEtran}
\usepackage{graphicx}
\ifCLASSINFOpdf
\else
\fi

\usepackage{url}

\begin{document}
%
\title{Neuroevolution-Based Inverse Reinforcement Learning}

\author{\IEEEauthorblockN{Karan K. Budhraja and Tim Oates}\\
\IEEEauthorblockA{Computer Science and Electrical Engineering\\
University of Maryland, Baltimore County\\
Baltimore, Maryland 21250, USA\\
Email: karanb1@umbc.edu, oates@cs.umbc.edu}}

\maketitle

\begin{abstract}
The problem of Learning from Demonstration is targeted at learning to perform tasks based on observed examples. One approach to Learning from Demonstration is Inverse Reinforcement Learning, in which actions are observed to infer rewards. This work combines a feature based state evaluation approach to Inverse Reinforcement Learning with neuroevolution, a paradigm for modifying neural networks based on their performance on a given task. Neural networks are used to learn from a demonstrated expert policy and are evolved to generate a policy similar to the demonstration. The algorithm is discussed and evaluated against competitive feature-based Inverse Reinforcement Learning approaches. At the cost of execution time, neural networks allow for non-linear combinations of features in state evaluations. These valuations may correspond to state value or state reward. This results in better correspondence to observed examples as opposed to using linear combinations. This work also extends existing work on Bayesian Non-Parametric Feature Construction for Inverse Reinforcement Learning by using non-linear combinations of intermediate data to improve performance. The algorithm is observed to be specifically suitable for a linearly solvable non-deterministic Markov Decision Processes in which multiple rewards are sparsely scattered in state space. A conclusive performance hierarchy between evaluated algorithms is presented.
\end{abstract}


%
\IEEEpeerreviewmaketitle

\section{Introduction}
\label{section:introduction}

\begin{figure*}[!t]
\centering
\includegraphics[width=\textwidth,height=60mm]{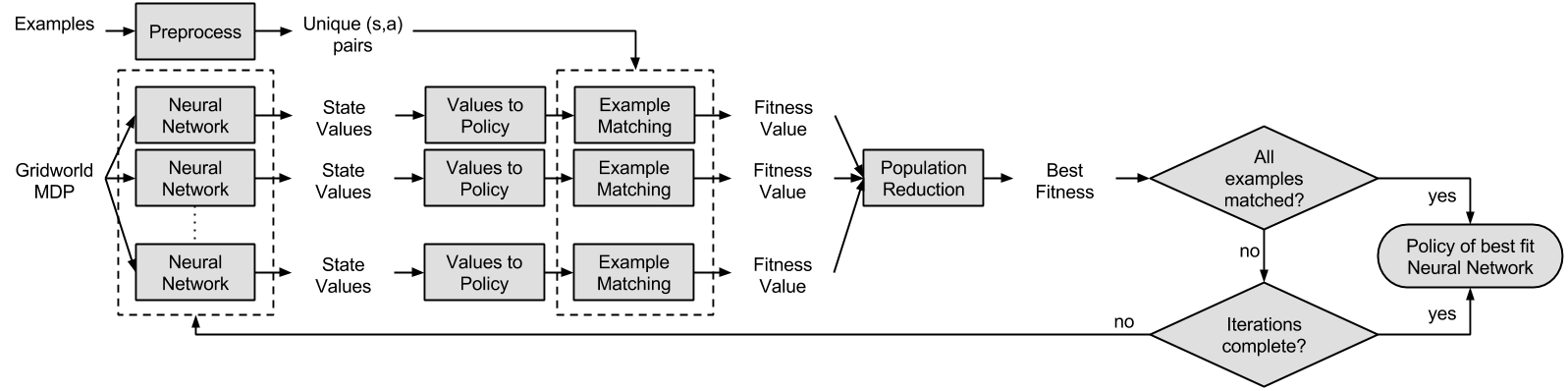}
\caption{NEAT-IRL summary. A population of neural networks is used to generate mappings from state features to state rewards. The networks are evolved in structure and connections using Genetic Algorithms (GA). Fitness of a network is determined by closeness of the optimal policy generated from rewards based on that network, to the demonstrated policy.}
\label{fig:neat_irl_summary}
\end{figure*}

The concept of Reinforcement Learning (RL) is motivated by modeling learning from experience. The environment is segmented into states, each of which contain information to describe the environment in that segment. The learner, also called the agent, may benefit differently depending on which state it is in. This creates a notion of rewards corresponding to each state. RL then seeks to find the optimal actions over all the states based on rewards  observed when taking actions in those states over time. An example would be a child learning how to build a pyramid from a set of blocks.

Inverse Reinforcement Learning (IRL) is motivated by learning from examples, such as someone showing a child how to build the block pyramid and the child then trying to replicate the process. As opposed to RL, an agent in IRL does not observe rewards; it attempts to recover them based on observed examples. The observations are in the form of traces of state-action pairs. IRL allows an agent to understand its environment in terms of evaluation of a state. It is therefore intuitive that IRL is a means to implement Learning from Demonstration (LfD) \cite{IRL2012,ng2000algorithms}. LfD describes a problem in which an agent learns to perform a task by observing how it is to be done. The learner exploits the demonstrator's experience and does not have to learn from experience itself.

One approach to IRL is to exploit state information by reconstructing rewards from state features and their combinations. To obtain such a non-linear function, this work employs a neural network as its first contribution. Specification of network structure can limit the efficiency of the generated network. For this reason, the neural network is generated by cumulative modification (using a Genetic Algorithm) to a simple feed-forward network. The approach is summarized in Figure \ref{fig:neat_irl_summary} with an overview in Section \ref{subsec:NEATIRL}.

As opposed to mapping states to rewards as in \cite{levine2010feature}, this work proposes generation of state values, from which state rewards can be derived. This improves robustness to noise, since state values typically do not exhibit steep transitions. Use of state values therefore favours real-world applications such as robotics.

The use of neural networks also allows for inherent advantages over regression trees \cite{caruana2006empirical,bengio2009learning}. Unlike regression trees (used in \cite{levine2010feature}), neural networks are capable of learning non-linear data boundaries. They are able to generate more abstract features at hidden neurons. Finally, the fact that neural networks can approximate any function with sufficient data (universal approximators) makes them intuitively preferable.

Further, when fitting a highly non-uniform function, neural networks are better than kernel functions (used in Gaussian Process (GP) regression \cite{levine2011nonlinear}) at generalizing non-locally and scaling to larger datasets. This is because kernel functions typically generalize locally.

Work in \cite{choi2013bayesian} involves the use of composite state features with priors to estimate rewards over several algorithm iterations. The second contribution of this work is utilizing the non-linear nature of neural networks to improve the performance of \cite{choi2013bayesian} in the case of complex reward structure. This is done by using a neural network for reward estimation.

Section \ref{section:relatedWork} describes relevant work on neuroevolution and feature based IRL. Section \ref{section:problemDefinition} discusses IRL and a feature based approach to the problem. This is followed by details of the proposed neuroevolution based algorithm in Section \ref{section:proposedMethod}. Section \ref{section:experiments} on experimental evaluation of the algorithm is followed by the concluding remarks in Section \ref{section:conclusion}.

\section{Related Work}
\label{section:relatedWork}

Feature construction for Inverse Reinforcement Learning (FIRL) uses regression trees and quadratic programming \cite{levine2010feature}. Optimization then involves selection of a sub-tree without significant loss in regression fitness. In \cite{levine2011nonlinear}, FIRL is acknowledged to have limited capabilites in representing a reward function that uses a non-linear combination of features. Another such technique, called GPIRL, is based on Gaussian Process (GP) regression \cite{levine2011nonlinear} and fits the reward function into a GP kernel as a non-linear combination of state features.

Recent techniques have also incorporated a non-parametric Bayesian framework to improve learning. This means that the number of parameters used by the models increases based on the training data. Composite features in \cite{choi2013bayesian} are defined as logical conjunctions of state features. The IRL model extends \cite{ramachandran2007bayesian} by defining a prior on these composite features. In \cite{michini2012bayesian} and \cite{michini2015bayesian}, the reward function is additionally assumed to be generated by a composition of sub-tasks to be performed in the MDP space. This algorithm targets detection of sub-goals but does not estimate the final policy over all states in state space (it only targets states observed in the demonstration).

Bayesian Non-Parametric FIRL (BNP-FIRL) \cite{choi2013bayesian} uses the Indian Buffet Process (IBP) \cite{ghahramani2005infinite} to define priors over composite state features. The IBP defines a distribution over infinitely large (in number of columns) binary matrices. It is used to determine the number of composite features and the composite features themselves. Features and corresponding weights are recorded over several iterations of the algorithm. These values are then either aggregated (as a mean-based result) or used for estimating Maximum A-Posteriori (MAP) \cite{choi2011map} (as a MAP-based result) state reward values. While the two results provide competitive performance when calculating reward functions, using the mean-based result performs significantly better than the MAP-based result when focusing on action matching rather than reward matching. For this reason, MAP-based results are excluded from comparison. More importantly, this emphasizes the importance of how the state reward values computed per iteration are finally used. A non-linear combination of this data intuitively provides better performance than a linear combination (as in the case of mean). Experimental results in \cite{choi2013bayesian} indicate superiority of BNP-FIRL over GPIRL in a non-deterministic (a certain percentage of actions are random) MDP, such as the ones used to evaluate our work.

In an expectation-maximization based approach, the reward function is modeled as a weighted sum of functions in \cite{hahn2015inverse}. Parameters of the optimal policy in the model are defined as the probability distribution of actions for each state in the state space, based on the optimal policy. The algorithm then attempts to simultaneously estimate the weights and parameters. The algorithm is not compared with \cite{levine2011nonlinear}, but the two algorithms have been individually compared with standard Maximum Entropy IRL. Visual observation of performance of these two algorithms indicates that a GP kernel based approach is competitive to, if not better than, an expectation-maximization based approach.

Very recently, the use of Deep Learning \cite{deng2014deep} for IRL problems has been explored in \cite{wulfmeier2015deep} and \cite{hewdeep}. The inputs to the first layer of the deep neural network are state features. The performance of this algorithm has been shown to surpass that of existing algorithms \cite{levine2011nonlinear,choi2013bayesian}. The algorithm focuses on achieving correct expected state value, whereas our algorithm focuses on learning the optimal policy. Intuitively, it is expected that our work will perform competitively with the use of deep neural networks. The reason for this is that the premise of both algorithms is similar: they use state features as input to a neural network and evaluate state reward or state value as the output of the neural network. In addition, the use of neuroevolution (evolving the structure of a neural network based on task requirements) allows for a more compact network due to dynamic construction of the network. Comparison with this technique is therefore regarded as outside of the scope of this work.

Finally, work in \cite{vroman2014maximum} on Maximum Likelihood IRL (MLIRL) covers three problem spaces: linear IRL, non-linear IRL and multiple intentions IRL. Linearity and non-linearity is in the context of the reward function modeled as a function of state features. Multiple intentions refers to an IRL setting where an MDP comprises of multiple reward functions. The algorithm emphasizes that other IRL methods are not suitable for a unified approach over all the mentioned problem spaces. However, it is noted that specialized IRL algorithms are more suitable if the number of experts and the reward function shape (linear or non-linear) is known. Its performance against other IRL algorithms is competitive. Performance of MLIRL for our problem setting is therefore evaluated as at most competitive with \cite{levine2011nonlinear} (evaluated in \cite{vroman2014maximum}). MLIRL is therefore excluded from comparison for our work.

On the lines of FIRL, a neural network can provide a mapping from state features to state value. Since a neural network can compactly represent complex combinations of inputs, internal neurons may represent more informative features, as those obtained in FIRL. In an alternative implementation extending BNP-FIRL, a neural network is used to provide state value based on feature and weight data gathered over several iterations of state reward estimation in the BNP-FIRL algorithm.

Since the function to be generated is unknown, so is its complexity. This implies uncertainty about the optimal number of layers and nodes in each layer to be used. Nodes in later layers of a neural network may be able to define a function for which it may take several nodes in the earlier layers of the neural network to define. There is therefore a trade-off between the number of hidden layers and the number of nodes in each layer. Because of this, using a fixed structure for the neural network in this scenario may be sub-optimal. Neuroevolution solves this problem by generating the optimal neural network using techniques such as Genetic Programming (GP) \cite{gruau1994neural}, Evolutionary Programming (EP) \cite{angeline1994evolutionary}, Simulated Annealing (SA) \cite{yao1997new}, Genetic Algorithms (GA) \cite{stanley2005real,stanley2002evolving}, Evolution Strategies (ES) \cite{kassahun2005efficient,siebel2007evolutionary}, Evolutionary Algorithms (EA) \cite{rempis2012evolving} and Memetic Algorithms (MA) \cite{sher2012handbook}.

In a $direct$ $encoding$ scheme, all neurons and connections in the neural network are explicitly specified by the genotype. In case of an $indirect$ $encoding$ scheme, these values are expressed implicitly by smaller parts of the genotype (encoding of the neural network). $Indirect$ $encoding$ is suitable for solving tasks which have a high degree of regularity (such as controlling the legs of a millipede like robot). Approximating a function is a problem that lacks regularity. A well established $direct$ $encoding$-based neuroevolution technique such as NeuroEvolution of Augmenting Topolgies (NEAT) \cite{stanley2002evolving,NEAT2014}, which evolves both the structure and parameters of the neural network, is therefore preferred for our work. 

NEAT evolves a population of neural networks governed by a GA \cite{koza1992genetic,banzhaf1998genetic}, and therefore by a fitness function. There currently exist many extensions of NEAT, including rtNEAT \cite{stanley2005real} (a real-time version of NEAT, which enforces perterbation to avoid stagnation of fitness level) and FS-NEAT \cite{whiteson2005automatic} (NEAT tailored to feature selection). 

In \cite{yong2006incorporating}, demonstration bias is introduced to NEAT-based agent learning. This is done by providing advice in the form of a rule-based grammar. This does not, however, provide an embedded undertanding of preference for a state in state space. This is further explored by mapping states to actions in \cite{karpov2011human} where different methods of demonstration are studied in a video game environment \cite{stanley2005evolving}. NEAT-based IRL is also implemented in a multiple agent setting as in \cite{miikkulainen2012multiagent} where groups of genomes share fitness information.

State values are used to generate a corresponding policy. To incorporate learning by demonstration, this policy is matched with the demonstration and the neural network is evolved thereof. Such directed neuroevolution can be considered as an extension of \cite{yong2006incorporating,karpov2011human} with better insight to evaluation of a state. For the purpose of this document, the proposed work is referred to as NEAT-IRL.

\section{Problem Definition}
\label{section:problemDefinition}

A formal definition of an MDP \cite{puterman2014markov} is repeated here for reference. An MDP is a set of states ($S$), actions ($A$) and transition probabilities ($\theta$) between states when an action is taken in a state. Additionally, each state-action pair corresponds to a reward ($R$). A discount factor ($\gamma$) is used while aggregating rewards corresponding to a trajectory of state-action pairs. A policy ($\pi$) describes a set of actions to be taken over the state space. The optimal policy ($\pi^{*}$), then, maximizes the expected discounted sum of rewards between two given states ($start$ and $goal$) in an $episodic$ task (which repetitively solves the same problem). Alternatively, in the case of a $continual$ task (which does not have terminal states), the optimal policy maximizes this sum of rewards over the lifetime of the learning agent. State value ($v$) is the expected return (sum of discounted $R$ values) when an arbitrary $\pi$ is followed, starting at that state. The concept of an MDP is extended to define a Linear MDP (LMDP): a linearly solvable MDP (using KL-divergence \cite{kullback1951information} or maximum entropy control) \cite{ziebart2010modeling}. An LMDP is defined by state costs ($q$) in correspondence to $S$ as an alternative to $R$ \cite{vroman2014maximum}. Passive dynamics ($p$) describes transition probabilities in the absence of control. Following a policy ($\pi$) as opposed to $p$ occurs at a cost of the KL divergence between $\pi$ and $p$. Such a cost makes the optimization problem convex, removing questions of local optima \cite{todorov2006linearly}. Additionally, an exponential transformation of the optimal $v$ function transforms the associated Bellaman Equation \cite{sutton1998reinforcement} (MDP solution) to a linear function. An optimal $v$ function corresponds to a $v$ function evaluated over $\pi^{*}$.

The goal of IRL is to learn a reward function for each state based on parts of a given policy (a demonstration). In a broader sense, the goal is to be able to generate a policy over a state space ($S$), which is correlated to what has been demonstrated. For the purpose of this work, the demonstrations received by the algorithm are assumed to be performed by an expert, meaning that they are assumed to be optimal. A demonstration ($D$) consists of numerous examples, each of which is a trace of the optimal policy through state space. These are represented in the form of sequences of state-action pairs ($s,a$).

One method to generate a policy is to generate state values for each state based on state features. It is assumed that a weighted combination of state features can provide a quantitative evaluation of a state (with a motivation similar to FIRL). The first problem, then, is to learn a mapping from state features to state values that produces a policy for which state-action pairs are consistent with the given examples.

Additionally, values of weights and features over several iterations of BNP-FIRL may be used in different ways to derive a state reward value. The second problem, then, is to learn a non-linear mapping from these values to state reward, which produces a policy consistent with the given examples (as described for the first problem).

The domain used for this work is a grid world Markov Decision Process (MDP), but the concept can be extended to other state spaces. The use of a grid world MDP is aligned with experiments in \cite{levine2010feature,michini2012bayesian,michini2015bayesian,vroman2014maximum,hahn2015inverse}. 
In case of a grid world MDP, an agent has $5$ possible actions: move up, move down, move left, move right, do nothing. In case of a deterministic MDP, the action taken is always the action selected. However, in a non-deterministic MDP, the action taken is sometimes random, irrespective of the action selected.

\section{Proposed Method}
\label{section:proposedMethod}

\begin{figure}
\centering
\includegraphics[height=60mm]{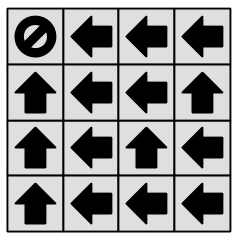}
\includegraphics[height=60mm]{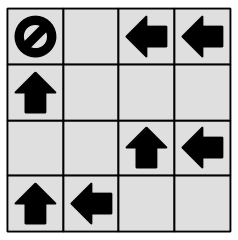}
\caption{Policy sampling in a grid world MDP: $4$ demonstrations of length $2$. The top part of the figure shows the optimal policy for the MDP. For each demonstration, a state is selected at random. The optimal policy is then followed for steps equal to demonstration length ($2$ in this case).}
\label{fig:policy_sampling}
\end{figure}

Examples are provided as traces of subsequent states in the state space as per the demonstration policy. Multiple traces may overlap, which means that a single state may be covered more than once in a set of examples. If no examples overlap, the set of states involved in the examples has an upper bound of the sum the length of each example (in states). An example from an MDP policy perspective is given in Figure \ref{fig:policy_sampling}. These examples serve as demonstration ($D$) in IRL (used for learning state rewards).

The following sections provide an overview of fundamental algorithms in the context of this work and then continue to define an approach to NEAT-based IRL.

\subsection{BNP-FIRL}

\begin{figure}
\centering
\includegraphics[width=2.5in]{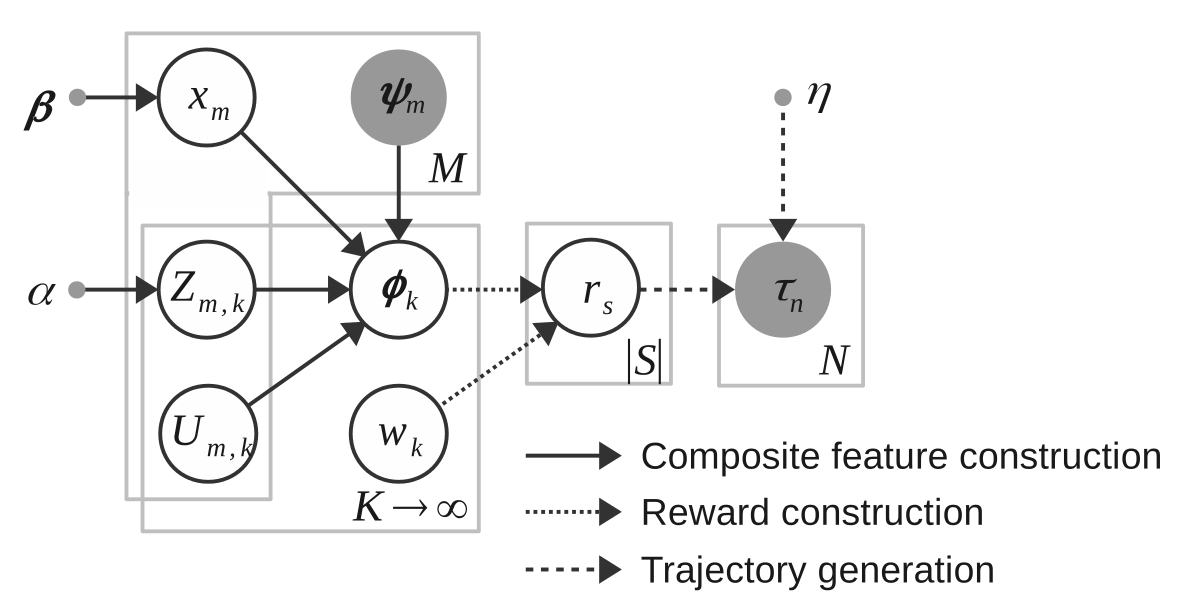}
\caption{BNP-FIRL summary \cite{choi2013bayesian}. The reward function $r$ is computed as a product of composite features $\Phi$ and associated weights $w$. Values of $\Phi$ and $w$ are selected to optimize the posterior probability of $\tau$ given $r$.}
\label{fig:bnpfirl_summary}
\end{figure}

BNP-FIRL is summarized in Figure \ref{fig:bnpfirl_summary}. BNP-FIRL decomposes reward ($r$) as a product of composite features ($\Phi$) and weights ($w$). There are a total of $K$ composite features and so the size of $w$ is $K$. Each column in $\Phi$ is a binary vector of size $|S|$, indicating the presence or absence of the feature per state. The size of $\Phi$ is therefore $K \times |S|$. The product of $w$ and $\Phi$ is thus a vector of size $|S|$.

$\Psi$ denotes a matrix of $M$ atomic features (the original state features as defined by the MDP). As for $\chi$, each column in $\chi$ contains $|S|$ binary data items, making its size $M \times |S|$.

The matrix $Z$ indicates the atomic features that comprise each composite feature. It is therefore $M \times K$ in size. A Bernoulli distributed binary matrix (with $p=0.5$), $U$ (also $M \times K$), is used to negate atomic features when generating composite features. $X$ is an $M$-dimensional binary vector indicating use of the atomic features to form composite features. 

$\alpha$ is a constant parameter over which a Poisson distribution is calculated (required by the IBP when computing the number and values of composite features). A Bernoulli distribution ($p=\kappa$) is used to generate priors on $X$. Finally, $\kappa$ is Beta distributed using the closed interval defined by $\beta$.

$\tau$ is a vector of $N$ demonstrations that is assumed to be generated using the optimal policy for the MDP. The posterior probability of $\tau$ given $r$ and constant $\eta$ is then computed. $\eta$ is the parameter representing the confidence of actions being optimal \cite{choi2013bayesian}. The algorithm iteratively converges to a value of $r$ which maximizes this probability. The values of $\Phi$ and $w$ computed in each iteration are stored in a vector. To compute mean-based results, the sum of rewards per iteration is used as the final reward. The use of mean at the final stage of the BNP-FIRL algorithm is denoted by BNP-FIRL(mean). 

\subsection{NEAT}

GAs \cite{koza1992genetic,banzhaf1998genetic} are biologically inspired population based stochastic search techniques. A population of $genomes$ is evolved in favour of optimizing a $fitness$ $function$. The performance of GA is primarily governed by two parameters: population size ($N_{P}$) and maximum number of generations ($N_{G}$).

An Artificial Neural Network (ANN) or a Neural Network (NN) \cite{haykin2004comprehensive} is a function approximation model inspired by biological neural networks. The model is composed of interconnected nodes or $neurons$ which exchange information to produce the desired output. In our work, we use neural networks containing a single output node.

\begin{figure*}
\centering
\includegraphics[width=\textwidth,height=60mm]{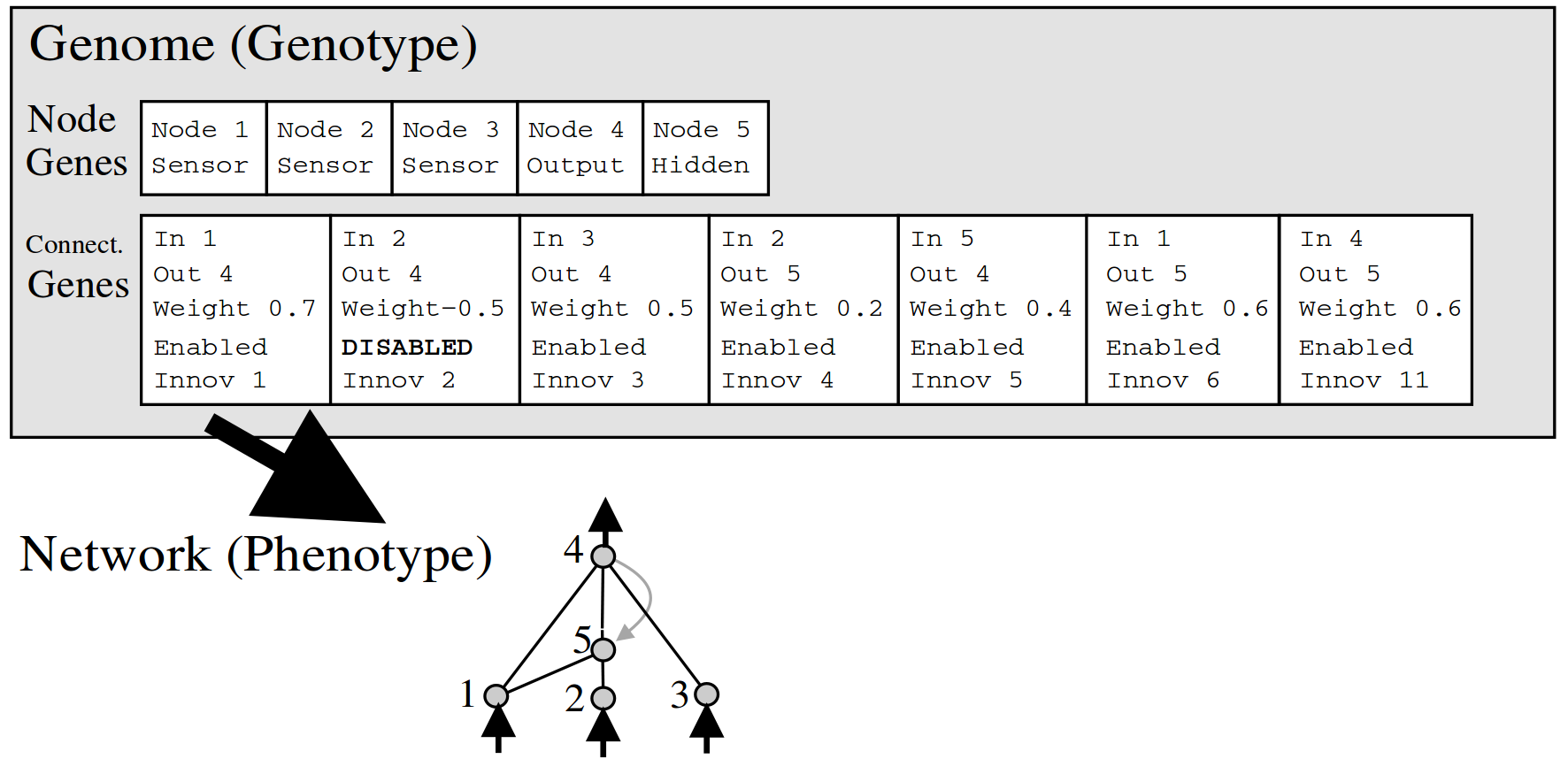}
\caption{Genotype to phenotype mapping example for NEAT \cite{stanley2002evolving}. A genotype corresponds to the encoding of the structure (nodes) and connections of a neural network. These are represented as two genes. The phenotype is the actual neural network generated based on its phenotype.}
\label{fig:genotype_phenotype}
\end{figure*}

NEAT evolves neural networks using GA, guided by a $fitness$ $function$. Each member of the population corresponds to a genotype or genome and a phenotype (the actual neural network). A genome consists of node genes (listing input, output and hidden nodes) and connection genes (listing connections between nodes and associated weights). Genotype to phenotype mapping is summarized in Figure \ref{fig:genotype_phenotype}. NEAT begins with relatively less complex neural networks and then increases complexity based on a $fitness$ requirement. Specifically, the initial network is perceptron-like and only comprises of input and output neurons. The gene is evolved by either addition of a neuron into a connection path or by creating a connection between existing neurons.

\subsection{NEAT-IRL}

The result of NEAT-IRL is a neural network which can produce state values based on state features.

Neural networks represented by a genome population in NEAT are considered to use state features as input and produce state value as output. A corresponding policy is evaluated for that state space. It is therefore suitable that the fitness function be the coherence between the generated and demonstrated policies. In implementation, this is done by using the coherence ($c$) of generated action directions ($d_{G}$) with the demonstration ($d_{D}$) as fitness value. 

\begin{equation}
c =
\left\{
	\begin{array}{ll}
		1  & \mbox{if } d_{G} = d_{D} $ (in correct direction)$\\
		-1 & \mbox{if } d_{G} = -d_{D} $ (opposite to the correct direction)$\\
		0 & \mbox{otherwise} $ (other wrong direction or no action)$\\
	\end{array}
\right.
\end{equation}

These values are intuitive to the cosine of the angle between the generated and example actions. Fitness is computed as $c$ accumulated over all states included in the demonstration. If all of the demonstrated actions are replicated correctly, the algorithm is terminated. This is summarized in Figure \ref{fig:neat_irl_summary}.

\subsection{BNP-FIRL(NEAT)}

The use of NEAT is similar to NEAT-IRL where the fitness value is based on matching examples in the demonstration. However, the inputs are no longer state features.

The dimensions of $\Phi$ and thus $w$ vary across iterations because of variation in composite features considered. A non-linear combination of $\Phi$ or $w$ over time is therefore not possible. However, the size of $r$ is constant for each iteration. The set of values of $r$ over iterations of the algorithm then serves as input to the neural network. The output of the neural network is used as the resultant state reward vector. This reward vector is used to compute the optimal policy, which can then be used to compute fitness values.

\subsection{FIRL and GPIRL vs NEAT-IRL}
\label{subsec:FIRLandGPIRLvsNEATIRL}

FIRL, GPIRL and NEAT-IRL all consider individual states as opposed to traces of state sequences, making them memory efficient in terms of the storage of examples. However, FIRL and GPIRL generate a function which produces state rewards, as opposed to NEAT-IRL which produces state values. State rewards evaluate $immediate$ desirability whereas state values evaluate long term desirability of a state. An agent therefore seeks actions that lead to states of higher value and not highest reward to maximize long term reward \cite{barto1998reinforcement}. State values also observe smoother transition amongst states in close proximity, though this is not necessary for state rewards. Further, converting a set of state values to a policy requires a single computational step (value based greedy action selection). In case of state rewards, however, the policy is accumulated over several $epochs$ of computation, thereby resulting in greater time complexity. Additionally, a policy generated using state values is more robust to noise (in context of real-world applications) than one generated based on state rewards. The reason is that in case of state values, policy evaluation only requires a comparison of adjacent state values. The final policy is therefore only affected if the state values are close enough for noise to change action selection. In case of policy generation using state rewards, noise would be accumulated over each $epoch$ of computation and would therefore have a more significant effect on the generated policy. A disadvantage of generating state values is that they transfer poorly to other MDPs with similar feature sets as opposed to state rewards \cite{vroman2014maximum}. However, our problem space does not concern transfer learning.

Additionally, FIRL involves convex optimization (minimization) and the use of regression trees, whereas NEAT-IRL is built on neural networks. 

\cite{lilley2005neural} shows that GP behaviour can be replicated by a multi layer perceptron neural network with a sufficiently (tending to infinity) large number of hidden neurons. This stands with a requirement to use weight decay \cite{neal1996priors,neal1995bayesian}. In practice, this implies limitation of neural networks for approximation of GP behaviour instead of exactness. Further, GP models can be optimized to fit data exactly with specific hyper-parameter values \cite{rasmussen2006gaussian}. This implies a trade-off between exactness and over-fitting data.

NEAT-IRL does, however, introduce two new parameters ($N_{P}$ and $N_{G}$) to the IRL problem, which increases degrees of freedom. The performance of a fixed set of parameters will vary in different environments. Algorithm performance may therefore need to be evaluated across these parameters for optimal value assignments.

\subsection{BNP-FIRL(mean) vs BNP-FIRL(NEAT)}

The set of functions defined by a linear combination of variables is a subset of the set of functions defined by a non-linear combination of those variables. Non-linear combinations are therefore more powerful in expressing relationships among variables and direct towards better function approximations at the cost of function complexity. In the case of the mean-based result of BNP-FIRL in particular, the linear combination is simply a sum of variables. This allows for significant scope for improvement in approximation of the final value of state rewards.

Similar to NEAT-IRL, the use of NEAT with BNP-FIRL increases the number of algorithm parameters and may require an additional level of optimization for better algorithm performance.

\section{Experiments}
\label{section:experiments}

\begin{figure}
\centering
\includegraphics[width=2.5in]{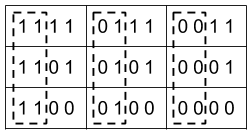}
\caption{State features for a gird world MDP ($n=3$). There are $2\times(3-1)=4$ features per state. Features have binary values and demonstrate patterns across rows and columns.}
\label{fig:state_features}
\end{figure}

Originally conceptualized in a Python implementation based on MultiNEAT \cite{MultiNEAT2012}, NEAT-IRL is currently implemented in MATLAB using \cite{MATLABNEAT2003} and is evaluated using existing tools in the IRL toolkit containing FIRL and GPIRL \cite{IRLToolkit2011}. The implementation of BNP-FIRL exists in an extended version of the toolkit \cite{IRLToolkit2013}. It is used with \cite{IRLToolkit2011} for the experiments in this work. Inherent to the toolkit, there exist $2 \times (n-1)$ binary state features for a grid of size $n$. These bit patterns contain sub patterns which are consistent for a row or column in the grid, thereby forming a coordinate system. This is exemplified in Figure \ref{fig:state_features}. Additionally, state rewards are assigned randomly for each macro block. The algorithms are evaluated over many such randomly generated grid worlds and are compared on how well they estimate the randomly generated reward structure.

FIRL and GPIRL consistently produce better results than the other IRL algorithms they are compared to in \cite{levine2010feature,levine2011nonlinear}. The case is similar with BNP-FIRL(mean) \cite{choi2013bayesian}. Additionally, GPIRL performs consistently better than FIRL. It is therefore sufficient to evaluate GPIRL, NEAT-IRL, BNP-FIRL(mean) and BNP-FIRL(NEAT) when examining performance improvements. NEAT-IRL is not compared with the work done in \cite{yong2006incorporating,karpov2011human} in sight of a dependency on rule based learning.

The algorithms are evaluated in a number of ways. NEAT-IRL is evaluated individually and also in comparison to GPIRL, BNP-FIRL(mean) and BNP-FIRL(NEAT). The IRL toolkit \cite{IRLToolkit2011} defines misprediction score as the probability that the agent will take a non-optimal action (different from what would have been in an example) in a state. This is measured based on matching the expert policy (used for demonstration) and the policy generated by the IRL algorithm. These scores are evaluated for both linear and standard MDPs. With $5$ possible actions at any state in a grid world, default misprediction score is $0.8$. Macroblock size, $b$, which specifies the number of adjacent cells in a grid to be assigned the same reward value, is set to $1$ for all experiments. This is done so that state features correspond to unique rewards. Average values are computed over $25$ executions. Furthermore, NEAT-IRL may end execution early when a generated policy completely matched with demonstrated examples. This may, however, lead to under-fitting, which is also a possible contributor to hindering the performance of NEAT-IRL in terms of misprediction score.

Note that the default values used for NEAT in \cite{MATLABNEAT2003} are $N_{P}=150$ and $N_{G}=200$. Additionally, GPIRL is evaluated in \cite{IRLToolkit2011} using a default grid size ($n$) of $32$, i.e. a $32\times32$ grid with $16$ training samples ($N_{S}$) of length ($L_{S}$) $8$ each. Reduced values (scaled in proportion) are used for evaluation of NEAT-IRL and BNP-FIRL(NEAT) for computational tractability. Primary results discussed in this work use the configuration $n=16$, $N_{S}=8$ and $L_{S}=4$. Certain supplementary evaluations use the configuration $n=4$, $N_{S}=4$ and $L_{S}=1$. The number of samples in this case is more than the scaled value ($N_{S}=2$) to avoid inference based on little data. In doing so, the demonstrations include $25\%$ of the total states, which is justified by considering the proportions used in the original setting ($n=32$). The primary grid allows for standardized evaluation of the algorithms, whereas a smaller grid allows more tractable analysis of the MDPs over which the algorithms are evaluated. The remaining NEAT configuration including GA parameters are as used in \cite{MATLABNEAT2003}. GPIRL parameters are used as default in \cite{IRLToolkit2011}. Note that for smaller data samples, interpolation used by GPIRL exceeded the possible number of points, thereby causing mathematical errors. This was fixed by limiting interpolation to the maximum possible, in case the number of interpolations was more than what was possible.

In the misprediction score graphs plotted, a solid line is used to denote performance data on a standard MDP and a dotted line is used to denote performance data on a linear MDP. Additionally, the term computational complexity is interchangeably used with computational time complexity and execution time is calculated in seconds. Additionally, misprediction score is observed to be lower when testing on a linear MDP than on a standard MDP irrespective of which of these two MDP types was used for training the algorithm.

\subsection{NEAT-IRL}
\label{subsec:NEATIRL}

Evaluation begins with testing NEAT-IRL parameters $N_{P}$ and $N_{G}$. This is done to evaluate the effect of NEAT parameters on algorithm performance and complexity.

As expected in the use of GAs, a larger population of genomes provides better performance. Note that beyond a threshold value, an increase in population does not contribute to significant optimization. This is because the capacity of search that can be performed by the extra (beyond threshold) population of genomes is limited by the size of the current state space. Execution time for a linear implementation of NEAT-IRL is linear with population size.

Coherent to GA behaviour, an increase in the number of maximum generations results in lower misprediction score. This is subject to stagnation in a manner similar to that discussed for population size. NEAT-IRL execution time is linear with the maximum number of generations as expected, since computation is constant per generation of the algorithm.

\subsection{GPIRL and BNP-FIRL(mean) vs NEAT-IRL and BNP-FIRL(NEAT)}

GPIRL, BNP-FIRL(mean), NEAT-IRL and BNP-FIRL(NEAT) are compared across three aspects of the MDP: the MDP type (standard or linear), the amount of determinism in the MDP ($d$, where a value of $1.0$ represents complete determinism) and different values of $N_{S}$. This tests the ability of the algorithms to reconstruct the reward function across different amounts of data. For each of these evaluations, NEAT parameters are arbitrarily set as $N_{P}=50$ and $N_{G}=50$.

\begin{figure}[!t]
\centering
\includegraphics[width=2.5in]{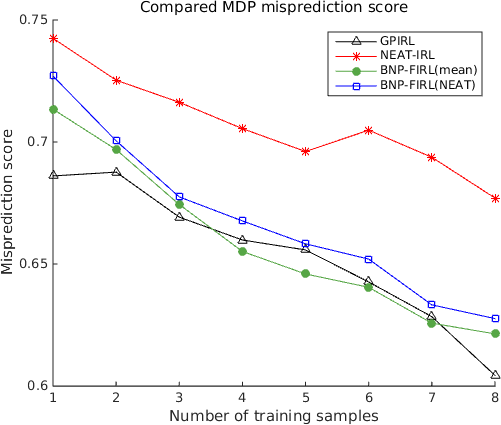}
\includegraphics[width=2.5in]{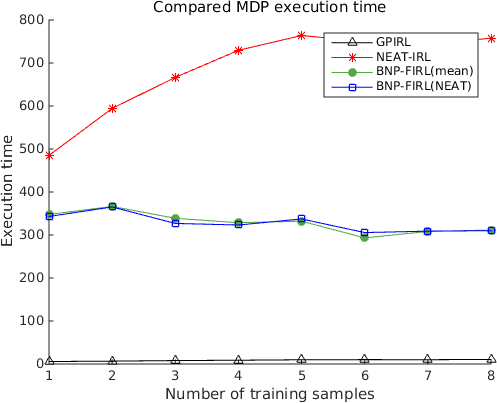}
\caption{Number of samples evaluation (standard MDP, $d=0.7$). The misprediction performance of BNP-FIRL(NEAT) is competitive with that of GPIRL and BNP-FIRL(mean). Time complexity is least for GPIRL, with that of BNP-FIRL(NEAT) being competitive with that of BNP-FIRL(mean).}
\label{fig:misprediction_standard_d07}
\end{figure}

\begin{figure}[!t]
\centering
\includegraphics[width=2.5in]{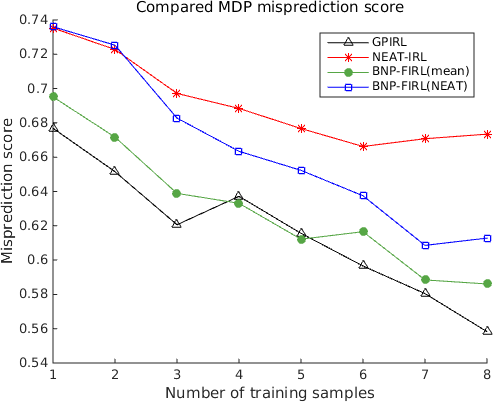}
\includegraphics[width=2.5in]{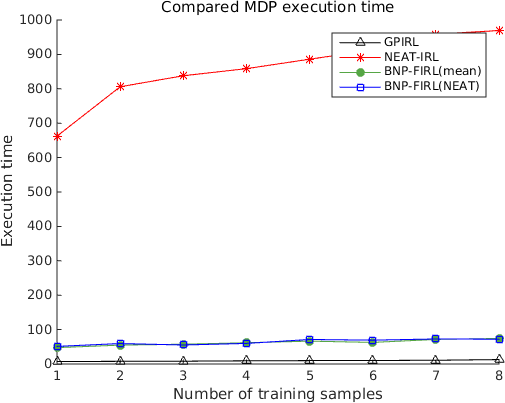}
\caption{Number of samples evaluation (standard MDP, $d=1.0$). GPIRL outperforms BNP-FIRL(NEAT), BNP-FIRL(mean) and NEAT-IRL on misprediction score. The time complexity for NEAT-IRL is significantly greater than that of the other IRL methods.}
\label{fig:misprediction_standard_d10}
\end{figure}

\begin{figure}[!t]
\centering
\includegraphics[width=2.5in]{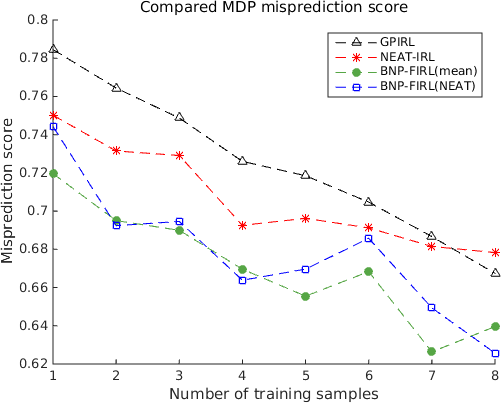}
\includegraphics[width=2.5in]{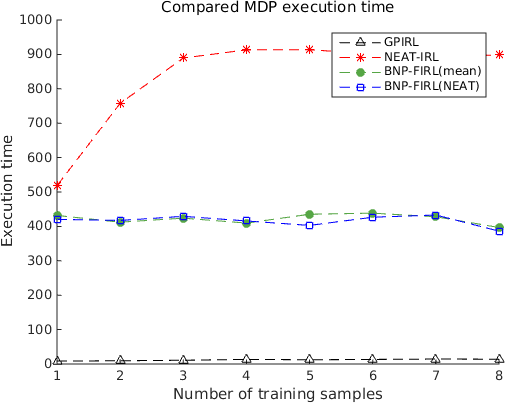}
\caption{Number of samples evaluation (linear MDP, $d=0.7$). The collective misprediction performances of BNP-FIRL(NEAT) and BNP-FIRL(mean) are better than better than those of GPIRL and NEAT-IRL. Time complexity is least for GPIRL, with that of BNP-FIRL(NEAT) being competitive with that of BNP-FIRL(mean).}
\label{fig:misprediction_linear_d07}
\end{figure}

\begin{figure}[!t]
\centering
\includegraphics[width=2.5in]{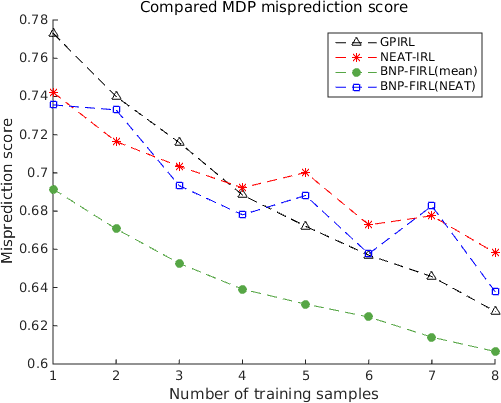}
\includegraphics[width=2.5in]{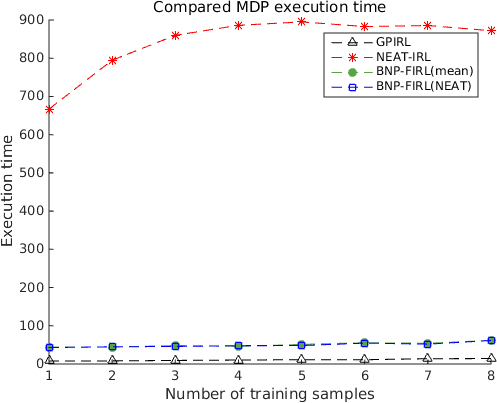}
\caption{Number of samples evaluation (linear MDP, $d=1.0$). BNP-FIRL(mean) outperforms BNP-FIRL(NEAT), GPIRL and NEAT-IRL on misprediction score. The time complexity for NEAT-IRL is significantly greater than that of the other IRL methods.}
\label{fig:misprediction_linear_d10}
\end{figure}

Dependency on $N_{S}$ is evaluated by varying $N_{S}$ from $1$ to $8$. To evaluate dependency on $d$, performance of the algorithms in two settings ($d=0.7$ as in \cite{choi2013bayesian} and $d=1.0$) are tested. Both of these settings are tested on standard and linear MDPs. These experiments are depicted in Figure \ref{fig:misprediction_standard_d07}, Figure \ref{fig:misprediction_standard_d10}, Figure \ref{fig:misprediction_linear_d07} and Figure \ref{fig:misprediction_linear_d10}.

In context of misprediction score, the performance of algorithms using neural networks is more competitive with the compared algorithms in a non-deterministic setting than in a deterministic setting. This favours use of the algorithm in a real world setting where non-determinism exists because of various sources of noise. It is also the case that better performance of neural networks is observed for a linear MDP than for a standard MDP. This is attributed to the neural network being able to perform better on a more easily solvable MDP given a set of parameters. Additionally, the composite features and thus reward function over iterations provide better performance of neural networks in BNP-FIRL(NEAT) as compared to the use of state features in NEAT-IRL. As in \cite{choi2013bayesian}, BNP-FIRL(mean) is observed to be consistently better than GPIRL.

Performance of NEAT-IRL improves at a slower pace than the other algorithms. This is attributed to the values of $N_{P}=50$ and $N_{G}=50$ being unoptimized for the MDPs being used. It is demonstrated in section \ref{subsec:NEATIRL} that these values are tunable to improve the performance of the algorithm.

Execution time is least for GPIRL in all four experimental settings. In a trade-off with better performance, an increased execution time is observed for BNP-FIRL(mean). This increase is more significant in the case of $d=0.7$. This implies that a linear MDP is a harder problem for NPB-FIRL(mean) to solve. The additional layer of NEAT at the final stage of BNP-FIRL does not introduce noticeable increase in execution time. Note that this is also because of the scale of values used, which results in lower resolution between the two graphs.

Execution time for NEAT-IRL remains largest across all of the experimental settings. In the presence of less demonstration data, the algorithm may often match all examples and discontinue evolving the network further. As the number of examples increases, fitting becomes more difficult and causes an increase in the number of generations. This explains the gradual increase in time complexity. Since there is a limit on the maximum number of generations that may be executed, the time complexity later stagnates. BNP-FIRL internally uses a matrix multiplication based technique to solve the MDP given state calculated rewards. However, NEAT-IRL processes each state based on state values of neighbouring state values. It is therefore possible that the additional conditional sequences result in significantly increased time complexity for NEAT-IRL as opposed to BNP-FIRL(NEAT). Perhaps, then, if we use neural networks to generate state reward instead of state value, we could integrate with the MDP solution method used by BNP-FIRL and reduce time complexity. However, mapping from state features to rewards is argued to decrease performance as compared to mapping from state features to state values (as discussed in Section \ref{subsec:FIRLandGPIRLvsNEATIRL}).

\begin{figure}[!t]
\centering
\includegraphics[width=2.5in]{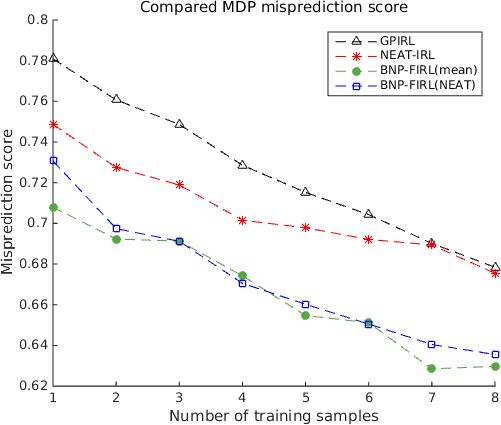}
\caption{Number of samples evaluation (linear MDP, $d=0.7$, $100$ executions). BNP-FIRL(mean) and BNP-FIRL(NEAT) collectively perform better than GPIRL and NEAT-IRL. An average over $100$ iterations asserts the competitive performances of FIRL(mean) and BNP-FIRL(NEAT).}
\label{fig:misprediction_linear_d07_100}
\end{figure}

\begin{figure}[!t]
\centering
\includegraphics[width=2.5in]{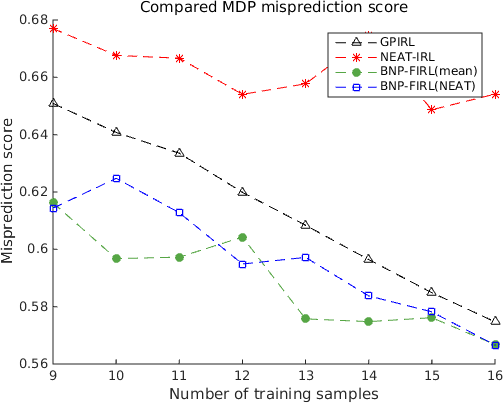}
\caption{Number of samples evaluation (linear MDP, $d=0.7$, $9-16$ samples). The performance of BNP-FIRL(NEAT) remains competitive with BNP-FIRL(mean). These two algorithms continue to perform better than GPIRL and NEAT-IRL even with increased number of training samples.}
\label{fig:misprediction_linear_d07_916}
\end{figure}

From a performance improvement perspective, Figure \ref{fig:misprediction_linear_d07} examines a setting where the use of neural networks provides competitive performance to the other algorithms in our examination set. However, a smoother graph is required for better comparison. For this reason, an average is considered over $100$ executions. The results are shown in Figure \ref{fig:misprediction_linear_d07_100}. Execution time data is similar on average to that in Figure \ref{fig:misprediction_linear_d07} and is therefore omitted. The figure establishes that the performance of BNP-FIRL(NEAT) is competitive to that of BNP-FIRL(mean), both of which are better than GPIRL and NEAT-IRL. The figure also establishes that in this setting, the use of neural networks alone (NEAT-IRL) outperforms GPIRL. The experiment in Figure \ref{fig:misprediction_linear_d07} is also extended to observe performance in the presence of $9-16$ samples (calculated over $25$ executions). This is shown in Figure \ref{fig:misprediction_linear_d07_916}. While the trends between GPIRL, BNP-FIRL(mean) and BNP-FIRL(NEAT) continue, limitation in performance of NEAT-IRL attributed to NEAT parameter settings is observed.

\begin{figure}
\centering
\includegraphics[height=60mm]{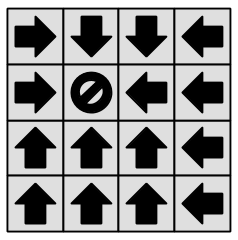}
\includegraphics[height=60mm]{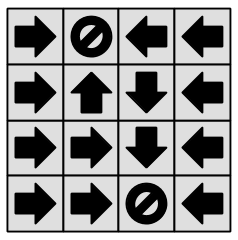}
\caption{MDP solutions (seeds $7$, $15$). The level of complexity of optimal policies is different for the two seed values. It is unclear whether the number of goal states (and therefore complexity of optimal policy) depicts a situation where BNP-FIRL(NEAT) would outperform BNP-FIRL(mean).}
\label{fig:mdp_solutions_7_15}
\end{figure}

\begin{figure}
\centering
\includegraphics[height=60mm]{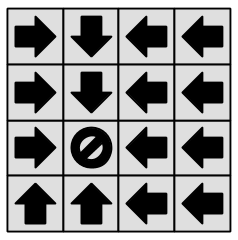}
\includegraphics[height=60mm]{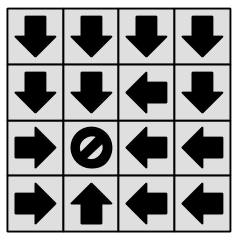}
\caption{MDP solutions (seeds $24$, $25$). The level complexity of optimal policies is low for both of the seed values. It is therefore unclear what part of the MDP impacts the performance of BNP-FIRL(mean) as opposed to BNP-FIRL(NEAT).}
\label{fig:mdp_solutions_24_25}
\end{figure}

To compare BNP-FIRL(mean) and BNP-FIRL(NEAT) algorithms, the use of neural networks (BNP-FIRL(NEAT)) is tested for its better adaptability to non linear boundaries as compared to BNP-FIRL(mean). The algorithms are evaluated over various non-deterministic ($d=0.7$) linear MDPs (varied by random seed initialization) to observe algorithm performance over different optimal policies for different MDPs. Overall, the performance of the two algorithms is indistinguishable. However, a closer look at specific seed values shows that BNP-FIRL(NEAT) performs noticeably better than NPBFIRL(mean) for certain $seed$ values. The policies for the MDPs for these situations are shown in Figure \ref{fig:mdp_solutions_7_15} and Figure \ref{fig:mdp_solutions_24_25}.

From MDPs corresponding to these seed values, it is unclear on whether the number of goal states discriminates performance between the two algorithms. It is then hypothesized that BNP-FIRL(NEAT) performs better than BNP-FIRL(mean) in the presence of multiple goal states. To evaluate this, two MDPs ($n=4$) are manually constructed with goals randomly placed in the MDP. This is done by associating an arbitrary reward value of $100$ to those states. The differences in performances of the two algorithms are also compared for significance using two tailed t-test. The results are summarized in Table \ref{tab:bnp_firl_goal_variation}. $M_{BNP-FIRL(mean)}$ and $M_{BNP-FIRL(NEAT)}$ represent misprediction scores corresponding to BNP-FIRL(mean) and BNP-FIRL(NEAT) respectively. To overcome the fluctuation of numbers for averages over smaller number of runs (such as $25$), the results are averaged over $1000$ runs.

\begin{figure}
\centering
\includegraphics[height=60mm]{1_goal.png}
\includegraphics[height=60mm]{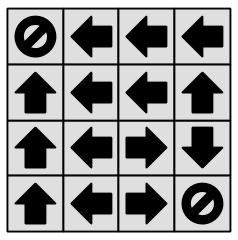}
\caption{Example MDP solutions ($1$ goal, $2$ goals). In the presence of less goals, the optimal policy shows smooth direction towards the goal(s). This can be encoded using a simple (more linear) reward function as opposed to a more non-linear function required in the presence of a more complex policy (as in Figure \ref{fig:mdp_solutions_3_4_goals}).}
\label{fig:mdp_solutions_1_2_goals}
\end{figure}

\begin{figure}
\centering
\includegraphics[height=60mm]{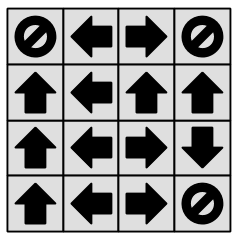}
\includegraphics[height=60mm]{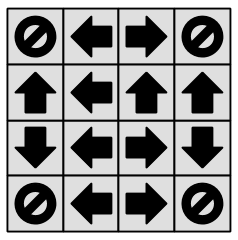}
\caption{Example MDP solutions ($3$ goal, $4$ goals). With an increase in the number of goals, the optimal policy becomes more complex than it is for less goals (as in Figure \ref{fig:mdp_solutions_1_2_goals}. The estimated reward function is therefore also expected to be more complex.}
\label{fig:mdp_solutions_3_4_goals}
\end{figure}

In the presence of a single goal state, BNP-FIRL(mean) significantly outperforms BNP-FIRL(NEAT). However, as the number of goals increases, BNP-FIRL(NEAT) eventually competes and later significantly outperforms BNP-FIRL(mean). From an MDP perspective, increasing the number of goals results in a more complex policy. This is because the state reward and hence state value surface has more than one optima. This is visually exemplified in Figure \ref{fig:mdp_solutions_1_2_goals} and Figure \ref{fig:mdp_solutions_3_4_goals}.

The hypothesis is then tested on a $n=16$ scale (with $N_{S}=16$, $L_{S}=4$ for correspondingly scaled input samples). $4$ goals with reward value $100$ are placed at each corner of the grid. In an average over $100$ executions, BNP-FIRL(NEAT) results in a significantly ($p-value=4.1819e-06$) lower misprediction score ($0.2672$) as compared to BNP-FIRL(mean) ($0.3072$). The hypothesis that the use of a neural network allows to learn more complicated reward structures is therefore confirmed. 

A conclusive performance hierarchy between these algorithms in a non-deterministic ($d=0.7$) linear MDP experimental setting is then established as BNP-FIRL(NEAT) $>$ BNP-FIRL(mean) $>$ NEAT-IRL $>$ GPIRL.

\begin{table}[!t]
  \centering
  \caption{Performance of BNP-FIRL(mean) and BNP-FIRL(NEAT) on constructed MDPs. BNP-FIRL(mean) performs better in the presence of less goal states. As the number of foal states increases and the reward function to be learned becomes more complex, BNP-FIRL(NEAT) begins to significantly outperform BNP-FIRL(mean).}
  \begin{tabular}{ l | l | l | l }
    \hline
    \# Goals & $BNP_{mean}$ & $BNP_{NEAT}$ & p-value \\ \hline
    \hline
	1 & 0.2308 & 0.2545 & 3.8837 e-4 \\
	2 & 0.3292 & 0.3392 & 0.1870 \\
	3 & 0.4119 & \bf{0.3913} & \bf{0.0063} \\
	4 & 0.4954 & \bf{0.4674} & \bf{3.9776 e-5} \\
    \hline
  \end{tabular}
  \label{tab:bnp_firl_goal_variation}
\end{table}

\section{Conclusion}
\label{section:conclusion}

These experiments conclude that algorithms using NEAT perform better on a non-deterministic linear MDP than BNP-FIRL(mean) and GPIRL (as in Figure \ref{fig:misprediction_linear_d07_100}). This is useful considering that real world MDPs contain uncertainty in action caused by various sources of noise.

Given the competitive performance of BNP-FIRL(NEAT) and BNP-FIRL(mean), hospitable MDPs for BNP-FIRL(NEAT) are then evaluated and experiments highlight the possibility of MDPs with multiple goals being favourable (as in Figure \ref{fig:mdp_solutions_7_15} and Figure \ref{fig:mdp_solutions_24_25}). This hypothesis is examined and superior performance of BNP-FIRL(NEAT) is confirmed in cases of MDPs containing multiple goal states. BNP-FIRL(NEAT) can to better estimate more complex reward structure (as in Table \ref{tab:bnp_firl_goal_variation}). A corresponding hierarchy of evaluated algorithms is then established with BNP-FIRL(NEAT) ranked the highest.

Additionally, NEAT parameters can be tuned to improve performance and time complexity for a given set of examples. The current implementation of NEAT-IRL is also capable of greater time efficiency. Computations specific to each genome in a population can be parallelized. Further, NEAT-IRL policy prediction is currently done for all states. This can be limited to only demonstrated states, since that is what determines fitness.

In future work, BNP-FIRL(NEAT) may be integrated to multiple agent settings and may also be extended to incorporate a cost of sharing information \cite{miikkulainen2012multiagent}.




%
\bibliographystyle{IEEEtran}
\bibliography{IEEEabrv,neat_irl}

\begin{thebibliography}{10}
\providecommand{\url}[1]{#1}
\csname url@samestyle\endcsname
\providecommand{\newblock}{\relax}
\providecommand{\bibinfo}[2]{#2}
\providecommand{\BIBentrySTDinterwordspacing}{\spaceskip=0pt\relax}
\providecommand{\BIBentryALTinterwordstretchfactor}{4}
\providecommand{\BIBentryALTinterwordspacing}{\spaceskip=\fontdimen2\font plus
\BIBentryALTinterwordstretchfactor\fontdimen3\font minus
  \fontdimen4\font\relax}
\providecommand{\BIBforeignlanguage}[2]{{%
\expandafter\ifx\csname l@#1\endcsname\relax
\typeout{** WARNING: IEEEtran.bst: No hyphenation pattern has been}%
\typeout{** loaded for the language `#1'. Using the pattern for}%
\typeout{** the default language instead.}%
\else
\language=\csname l@#1\endcsname
\fi
#2}}
\providecommand{\BIBdecl}{\relax}
\BIBdecl

\bibitem{IRL2012}
P.~Abbeel, ``Inverse reinforcement learning,''
  \url{http://www.cs.berkeley.edu/~pabbeel/cs287-fa12/slides/inverseRL.pdf},
  2012.

\bibitem{ng2000algorithms}
A.~Y. Ng, S.~J. Russell \emph{et~al.}, ``Algorithms for inverse reinforcement
  learning.'' in \emph{Icml}, 2000, pp. 663--670.

\bibitem{levine2010feature}
S.~Levine, Z.~Popovic, and V.~Koltun, ``Feature construction for inverse
  reinforcement learning,'' in \emph{Advances in Neural Information Processing
  Systems}, 2010, pp. 1342--1350.

\bibitem{caruana2006empirical}
R.~Caruana and A.~Niculescu-Mizil, ``An empirical comparison of supervised
  learning algorithms,'' in \emph{Proceedings of the 23rd international
  conference on Machine learning}.\hskip 1em plus 0.5em minus 0.4em\relax ACM,
  2006, pp. 161--168.

\bibitem{bengio2009learning}
Y.~Bengio, ``Learning deep architectures for ai,'' \emph{Foundations and
  trends{\textregistered} in Machine Learning}, vol.~2, no.~1, pp. 1--127,
  2009.

\bibitem{levine2011nonlinear}
S.~Levine, Z.~Popovic, and V.~Koltun, ``Nonlinear inverse reinforcement
  learning with gaussian processes,'' in \emph{Advances in Neural Information
  Processing Systems}, 2011, pp. 19--27.

\bibitem{choi2013bayesian}
J.~Choi and K.-E. Kim, ``Bayesian nonparametric feature construction for
  inverse reinforcement learning,'' in \emph{Proceedings of the Twenty-Third
  international joint conference on Artificial Intelligence}.\hskip 1em plus
  0.5em minus 0.4em\relax AAAI Press, 2013, pp. 1287--1293.

\bibitem{ramachandran2007bayesian}
D.~Ramachandran and E.~Amir, ``Bayesian inverse reinforcement learning,''
  \emph{Urbana}, vol.~51, p. 61801, 2007.

\bibitem{michini2012bayesian}
B.~Michini and J.~P. How, ``Bayesian nonparametric inverse reinforcement
  learning,'' in \emph{Machine Learning and Knowledge Discovery in
  Databases}.\hskip 1em plus 0.5em minus 0.4em\relax Springer, 2012, pp.
  148--163.

\bibitem{michini2015bayesian}
B.~Michini, T.~J. Walsh, A.-A. Agha-Mohammadi, and J.~P. How, ``Bayesian
  nonparametric reward learning from demonstration,'' \emph{Robotics, IEEE
  Transactions on}, vol.~31, no.~2, pp. 369--386, 2015.

\bibitem{ghahramani2005infinite}
Z.~Ghahramani and T.~L. Griffiths, ``Infinite latent feature models and the
  indian buffet process,'' in \emph{Advances in neural information processing
  systems}, 2005, pp. 475--482.

\bibitem{choi2011map}
J.~Choi and K.-E. Kim, ``Map inference for bayesian inverse reinforcement
  learning,'' in \emph{Advances in Neural Information Processing Systems},
  2011, pp. 1989--1997.

\bibitem{hahn2015inverse}
J.~Hahn and A.~M. Zoubir, ``Inverse reinforcement learning using expectation
  maximization in mixture models,'' in \emph{Acoustics, Speech and Signal
  Processing (ICASSP), 2015 IEEE International Conference on}.\hskip 1em plus
  0.5em minus 0.4em\relax IEEE, 2015, pp. 3721--3725.

\bibitem{deng2014deep}
L.~Deng and D.~Yu, ``Deep learning: methods and applications,''
  \emph{Foundations and Trends in Signal Processing}, vol.~7, no. 3--4, pp.
  197--387, 2014.

\bibitem{wulfmeier2015deep}
M.~Wulfmeier, P.~Ondruska, and I.~Posner, ``Deep inverse reinforcement
  learning,'' \emph{arXiv preprint arXiv:1507.04888}, 2015.

\bibitem{hewdeep}
M.~hew Alger, ``Deep inverse reinforcement learning.''

\bibitem{vroman2014maximum}
M.~C. Vroman, ``Maximum likelihood inverse reinforcement learning,'' Ph.D.
  dissertation, Rutgers University-Graduate School-New Brunswick, 2014.

\bibitem{gruau1994neural}
F.~Gruau \emph{et~al.}, ``Neural network synthesis using cellular encoding and
  the genetic algorithm.'' 1994.

\bibitem{angeline1994evolutionary}
P.~J. Angeline, G.~M. Saunders, and J.~B. Pollack, ``An evolutionary algorithm
  that constructs recurrent neural networks,'' \emph{Neural Networks, IEEE
  Transactions on}, vol.~5, no.~1, pp. 54--65, 1994.

\bibitem{yao1997new}
X.~Yao and Y.~Liu, ``A new evolutionary system for evolving artificial neural
  networks,'' \emph{Neural Networks, IEEE Transactions on}, vol.~8, no.~3, pp.
  694--713, 1997.

\bibitem{stanley2005real}
K.~O. Stanley, B.~D. Bryant, and R.~Miikkulainen, ``Real-time neuroevolution in
  the nero video game,'' \emph{Evolutionary Computation, IEEE Transactions on},
  vol.~9, no.~6, pp. 653--668, 2005.

\bibitem{stanley2002evolving}
K.~O. Stanley and R.~Miikkulainen, ``Evolving neural networks through
  augmenting topologies,'' \emph{Evolutionary computation}, vol.~10, no.~2, pp.
  99--127, 2002.

\bibitem{kassahun2005efficient}
Y.~Kassahun and G.~Sommer, ``Efficient reinforcement learning through
  evolutionary acquisition of neural topologies.'' in \emph{ESANN}, 2005, pp.
  259--266.

\bibitem{siebel2007evolutionary}
N.~T. Siebel and G.~Sommer, ``Evolutionary reinforcement learning of artificial
  neural networks,'' \emph{International Journal of Hybrid Intelligent
  Systems}, vol.~4, no.~3, pp. 171--183, 2007.

\bibitem{rempis2012evolving}
C.~W. Rempis, ``Evolving complex neuro-controllers with interactively
  constrained neuro-evolution,'' 2012.

\bibitem{sher2012handbook}
G.~I. Sher, \emph{Handbook of neuroevolution through Erlang}.\hskip 1em plus
  0.5em minus 0.4em\relax Springer Science \& Business Media, 2012.

\bibitem{NEAT2014}
K.~Stanley, ``The neuroevolution of augmenting topologies (neat) users page,''
  \url{http://www.cs.ucf.edu/~kstanley/neat.html}, 2014.

\bibitem{koza1992genetic}
J.~R. Koza, \emph{Genetic programming: on the programming of computers by means
  of natural selection}.\hskip 1em plus 0.5em minus 0.4em\relax MIT press,
  1992, vol.~1.

\bibitem{banzhaf1998genetic}
W.~Banzhaf, P.~Nordin, R.~E. Keller, and F.~D. Francone, \emph{Genetic
  programming: an introduction}.\hskip 1em plus 0.5em minus 0.4em\relax Morgan
  Kaufmann San Francisco, 1998, vol.~1.

\bibitem{whiteson2005automatic}
S.~Whiteson, P.~Stone, K.~O. Stanley, R.~Miikkulainen, and N.~Kohl, ``Automatic
  feature selection in neuroevolution,'' in \emph{Proceedings of the 7th annual
  conference on Genetic and evolutionary computation}.\hskip 1em plus 0.5em
  minus 0.4em\relax ACM, 2005, pp. 1225--1232.

\bibitem{yong2006incorporating}
C.~H. Yong, K.~O. Stanley, R.~Miikkulainen, and I.~Karpov, ``Incorporating
  advice into neuroevolution of adaptive agents.'' in \emph{AIIDE}, 2006, pp.
  98--104.

\bibitem{karpov2011human}
I.~V. Karpov, V.~K. Valsalam, and R.~Miikkulainen, ``Human-assisted
  neuroevolution through shaping, advice and examples,'' in \emph{Proceedings
  of the 13th annual conference on Genetic and evolutionary computation}.\hskip
  1em plus 0.5em minus 0.4em\relax ACM, 2011, pp. 371--378.

\bibitem{stanley2005evolving}
K.~O. Stanley, B.~D. Bryant, and R.~Miikkulainen, ``Evolving neural network
  agents in the nero video game,'' \emph{Proceedings of the IEEE}, pp.
  182--189, 2005.

\bibitem{miikkulainen2012multiagent}
R.~Miikkulainen, E.~Feasley, L.~Johnson, I.~Karpov, P.~Rajagopalan, A.~Rawal,
  and W.~Tansey, ``Multiagent learning through neuroevolution,'' in
  \emph{Advances in Computational Intelligence}.\hskip 1em plus 0.5em minus
  0.4em\relax Springer, 2012, pp. 24--46.

\bibitem{puterman2014markov}
M.~L. Puterman, \emph{Markov decision processes: discrete stochastic dynamic
  programming}.\hskip 1em plus 0.5em minus 0.4em\relax John Wiley \& Sons,
  2014.

\bibitem{kullback1951information}
S.~Kullback and R.~A. Leibler, ``On information and sufficiency,'' \emph{The
  annals of mathematical statistics}, pp. 79--86, 1951.

\bibitem{ziebart2010modeling}
B.~D. Ziebart, ``Modeling purposeful adaptive behavior with the principle of
  maximum causal entropy,'' 2010.

\bibitem{todorov2006linearly}
E.~Todorov, ``Linearly-solvable markov decision problems,'' in \emph{Advances
  in neural information processing systems}, 2006, pp. 1369--1376.

\bibitem{sutton1998reinforcement}
R.~S. Sutton and A.~G. Barto, \emph{Reinforcement learning: An
  introduction}.\hskip 1em plus 0.5em minus 0.4em\relax MIT press Cambridge,
  1998, vol.~1, no.~1.

\bibitem{haykin2004comprehensive}
S.~Haykin and N.~Network, ``A comprehensive foundation,'' \emph{Neural
  Networks}, vol.~2, no. 2004, 2004.

\bibitem{barto1998reinforcement}
A.~G. Barto, \emph{Reinforcement learning: An introduction}.\hskip 1em plus
  0.5em minus 0.4em\relax MIT press, 1998.

\bibitem{lilley2005neural}
M.~Lilley and M.~Frean, ``Neural networks: a replacement for gaussian
  processes?'' in \emph{Intelligent Data Engineering and Automated
  Learning-IDEAL 2005}.\hskip 1em plus 0.5em minus 0.4em\relax Springer, 2005,
  pp. 195--202.

\bibitem{neal1996priors}
R.~M. Neal, ``Priors for infinite networks,'' in \emph{Bayesian Learning for
  Neural Networks}.\hskip 1em plus 0.5em minus 0.4em\relax Springer, 1996, pp.
  29--53.

\bibitem{neal1995bayesian}
R.~M. Neal, ``Bayesian learning for neural networks,'' Ph.D. dissertation,
  University of Toronto, 1995.

\bibitem{rasmussen2006gaussian}
C.~E. Rasmussen, ``Gaussian processes for machine learning,'' 2006.

\bibitem{MultiNEAT2012}
P.~Chervenski, ``Multineat,'' \url{http://multineat.com/}, 2012.

\bibitem{MATLABNEAT2003}
C.~Mayr, ``Matlab neat,'' \url{http://nn.cs.utexas.edu/?neatmatlab}, 2003.

\bibitem{IRLToolkit2011}
S.~Levine, Z.~Popovic, and V.~Koltun, ``Irl toolkit,''
  \url{http://graphics.stanford.edu/projects/gpirl/irl_toolkit.zip}, 2011.

\bibitem{IRLToolkit2013}
J.~Choi and K.-E. Kim, ``Irl toolkit,''
  \url{http://ailab.kaist.ac.kr/codes/bayesian-nonparametric-feature-construction-for-irl},
  2013.

\end{thebibliography}

%






\end{document}